\documentclass[runningheads]{llncs}
\usepackage{graphicx}
\usepackage{comment}
\usepackage{amsmath,amssymb} 
\usepackage{color}

\usepackage{algorithm,algorithmic}
\usepackage{multirow}
\usepackage{subfigure}
\usepackage{verbatim}

\begin{document}
\pagestyle{headings}
\mainmatter
\def\ECCVSubNumber{4213}  

\title{Siamese Keypoint Prediction Network\\
for Visual Object Tracking} 

\titlerunning{~}
%
\author{Qiang Li \and
Zekui Qin \and
Wenbo Zhang \and
Wen Zheng}
\authorrunning{Q. Li et al.}
%
\institute{Kuaishou Technology, Beijing, China
\email{\{liqiang03,qinzekui03,zhangwenbo,zhengwen\}@kuaishou.com}}
\maketitle

\begin{abstract}
Visual object tracking aims to estimate the location of an arbitrary target in a video sequence given its initial bounding box.
By utilizing offline feature learning, the siamese paradigm has recently been the leading framework for high performance tracking.
However, current existing siamese trackers either heavily rely on complicated anchor-based detection networks or lack the ability to resist to distractors.
In this paper, we propose the Siamese keypoint prediction network (SiamKPN) to address these challenges.
Upon a Siamese backbone for feature embedding, SiamKPN benefits from a cascade heatmap strategy for coarse-to-fine prediction modeling.
In particular, the strategy is implemented by sequentially shrinking the coverage of the label heatmap along the cascade to apply loose-to-strict intermediate supervisions.
During inference, we find the predicted heatmaps of successive stages to be gradually concentrated to the target and reduced to the distractors.
SiamKPN performs well against state-of-the-art trackers for visual object tracking on four benchmark datasets including OTB-100, VOT2018, LaSOT and GOT-10k, while running at real-time speed.
\footnote{The code is available at \url{https://github.com/ZekuiQin/SiamKPN}}
\end{abstract}

\section{Introduction}
\begin{figure}[t]
\begin{center}
\label{FigIntroDemo1}
  \includegraphics[width=.6\textwidth]{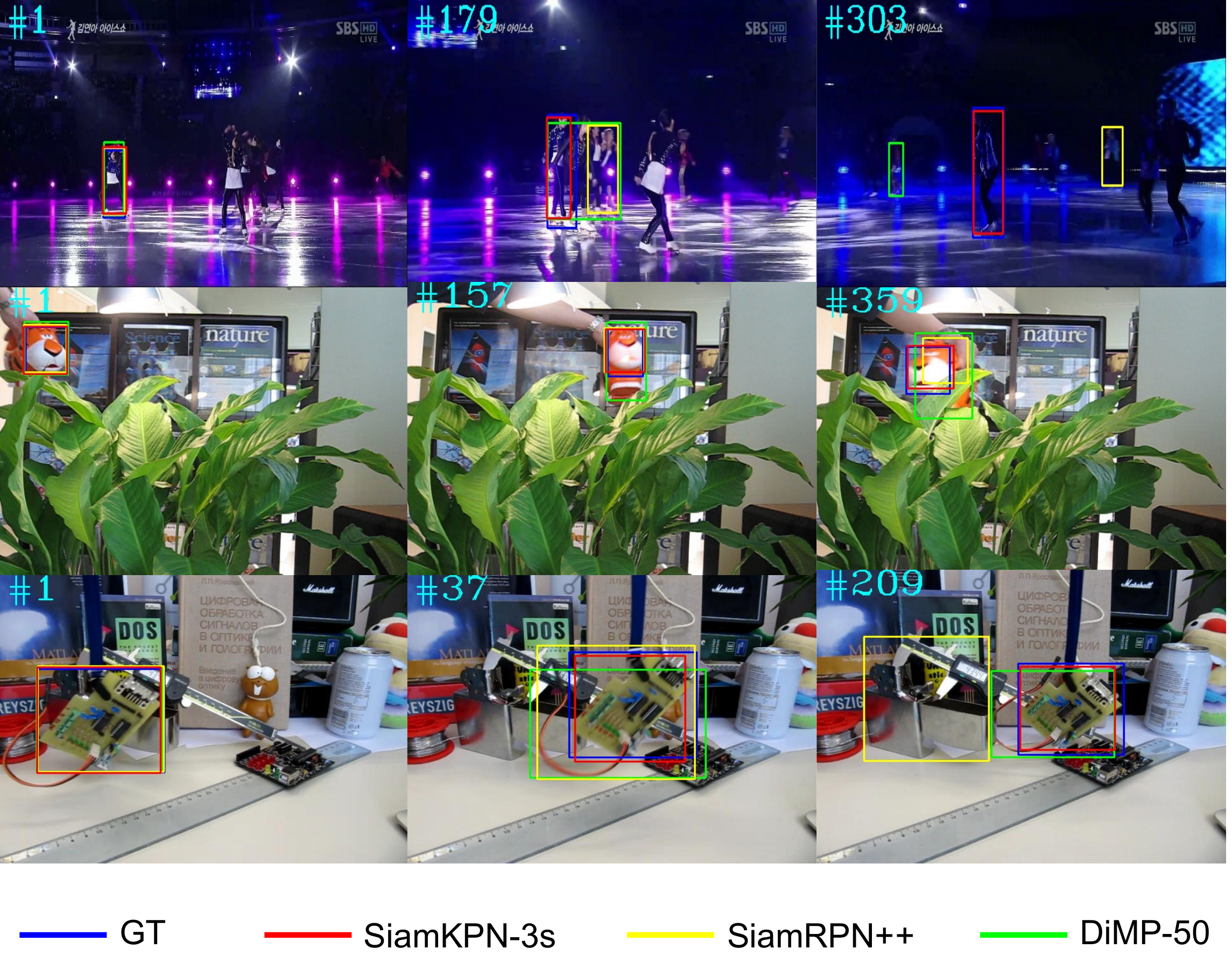}
\end{center}
\caption{Comparison among SiamRPN++ \cite{Li_2019_CVPR_SiamRPNpp},
DiMP-50 \cite{Bhat_2019_ICCV_DiMP} and SiamKPN-3s on three challenging sequences: Skating1, Tiger2 and Board.
Note that SiamKPN-3s can distinguish the target from distractors while the other methods drift to the background in Skating1.
In addition, SiamKPN-3s can better resist partial occlusion and scale variance to some extent.}
\end{figure}

Visual object tracking is the task of predicting the location of an arbitrary target in a video sequence provided only the target's bounding box in the first frame.
Like other computer vision tasks such as image classification, object detection and semantic segmentation, object tracking is very challenging due to appearance variations caused by deformation, viewpoint, scale, occlusion, illumination, etc.
Besides, the task is even more difficult to solve when considering the background clutter and similar distractors.
In literature, a classical approach to visual object tracking is the discriminative correlation filter \cite{bolme2010visual_MOSSE,henriques2014high_KCF}.
In the deep learning era, however, this approach is difficult to leverage end-to-end deep feature learning for even better performance.

To address this issue, the Siamese learning paradigm \cite{bromley1994signature} has been adopted and developed to harvest from offline deep feature learning \cite{Tao_2016_CVPR_SINT,bertinetto2016fully_SiamFC}.
This paradigm learns a shared feature embedding network for both the target region and the search image, thus formulates visual object tracking as a similarity learning problem.
The first implementation is SINT \cite{Tao_2016_CVPR_SINT} which trains the Siamese network by sampling pairs of patches.
although effective in tracking accuracy, SINT is far from real-time due to the redundant and inefficient patch-based feature extraction.
As a simple modification, SiamFC \cite{bertinetto2016fully_SiamFC} utilizes fully-convolutional operations to learn the Siamese network and solves the matching problem via an efficient cross-correlation between two feature maps.
Along this line of research, CFNet \cite{Valmadre_2017_CVPR_CFNet_SiamFCv2} incorporates the correlation filter as a differentiable layer into the Siamese framework.
RASNet \cite{Wang_2018_CVPR_RASNet} adopts the attention mechanism to improve the quality of the cross-correlation response.
In these methods, the prediction modeling part is restricted to simple architectures which results in limited performance to some extent.

To improve the prediction modeling, SiamRPN \cite{Li_2018_CVPR_SiamRPN} introduces the region proposal network (RPN) from the object detection literature \cite{ren2015faster} upon the Siamese network.
DaSiamRPN \cite{Zhu_2018_ECCV_DaSiamRPN} further handles distractors by augmenting the training data to include semantically hard negative pairs.
Recently, the backbone is successfully replaced by ResNet \cite{He_2016_CVPR_ResNet} in SiamRPN++ \cite{Li_2019_CVPR_SiamRPNpp} via a spatial aware sampling strategy to maintain useful translation invariance and the depth-wise cross-correlation to stablize training.
Another work \cite{Zhang_2019_CVPR_CIR} resorts to the cropping-inside residual (CIR) unit to modify ResNet, Inception \cite{szegedy2015going_Inception} and ResNeXt \cite{xie2017aggregated_ResNeXt} to accomodate deeper and wider backbone networks.

Concurrently, some methods utilize the cascade and the branching strategies to improve tracking accuracy and robustness of SiamRPN.
For example, C-RPN \cite{Fan_2019_CVPR_C-RPN} presents a Siamese cascade RPN framework to lend strength from stacking multiple RPNs which are coarse-to-finely trained according to adjusted anchor schemes.
SPM-Tracker \cite{Wang_2019_CVPR_SPM-Tracker} designs a series-parallel structure to fuse a coarse matching stage for robustness to distractors and a fine matching stage for discrimination power.
Though effective at large, all the above RPN-based methods heavily depends on the choice of the intricate anchor scheme to achieve reasonable tracking performance.


In this paper, we propose a Siamese keypoint prediction network (SiamKPN) for visual object tracking.
The whole network consists of a modified ResNet-50 Siamese backbone for feature learning, and a cascade of compact KPN heads for prediction modeling.
In particular, we employ the outputs of multiple layers from the backbone as features, while each KPN head is constructed by standard convolutions and one depth-wise cross-correlation.
By enforcing loose-to-strict intermediate supervisions, the cascade of predicted heatmaps can gradually concentrate to the target and reduce to the distractors.
Worth noting that, our method is motivated by both the cascade and the anchor-free strategy in the recent object detection literature \cite{Cai_2018_CVPR_CascadeRCNN,Law_2018_ECCV_CornerNet,Zhou_2019_arxiv_CenterNet}, though there are key differences.
Firstly, as far as our knowledge, we are the first to consider the anchor-free scheme in the Siamese paradigm for object tracking.
Secondly, the anchor-based cascade method adjusts the anchors for each stage to refine the prediction, while our novel cascade heatmap strategy applies loose-to-strict intermediate supervisions to guide the refinement process.

After offline training with a multi-task loss, SiamKPN provides an effective and efficient tracker without online updating.
Specifically, the SiamKPN tracker is robust to similar distractors to some extent, while running at real-time speed.
We evaluate the SiamKPN tracker through comprehensive experiments on four tracking benchmarks consisting of OTB-100 \cite{wu2015otb}, VOT2018 \cite{kristan2018vot}, LaSOT \cite{fan2019lasot} and GOT-10k \cite{huang2018got}.
In particular, the SiamKPN tracker with three stages (SiamKPN-3s) performs well against state-of-the-art deep trackers especially when compared with other Siamese trackers.
Figure \ref{FigIntroDemo1} presents some of the representative results on three challenging sequences.

\section{Related Works}
Aside from the Siamese tracking approach, we briefly review two other main categories of deep visual tracking methods considering the taxonomy of recent surveys \cite{li2018deep_survey,smeulders2013visual_survey,yilmaz2006object_survey}.
The two categories include feature-extraction tracking and end-to-end tracking.

\subsection{Feature-Extraction Tracking}
Most of the early deep tracking methods utilize deep networks just for feature extraction and rely on classical approaches for target prediction.
For example, CNN-SVM \cite{hong2015online_CNN-SVM} trains a support vector machine to classify positive and negative samples using the network outputs as appearance features.
This kind of region-based classification has to resort to sparse sampling for running speed while at the price of declined performance.

A better scheme is to train correlation filters based on the deep features.
For example, HCFT \cite{Ma_2015_ICCV_HCFT_CF2} adaptively learns correlation filters on multiple convolutional layers to encode the target appearance.
C-COT \cite{danelljan2016beyond_CCOT} presents a joint learning framework to fuse deep features from different spatial pyramids.
ECO \cite{Danelljan_2017_CVPR_ECO} introduces a factorized convolution operator, a generative sample space model and a conservative model update strategy to better utilize deep features for robust and efficient tracking.
UPDT \cite{Bhat_2018_ECCV_UPDT} proposes an adaptive fusion approach to leverage deep and shallow features to improve tracking performance.

\subsection{End-to-End Tracking}
End-to-end tracking usually learns a unified network to conduct both feature extraction and target prediction.
For example, DeepTrack \cite{li2015deeptrack} trains a simple CNN model of two convolutions and two fully-connected layers in a purely online manner for visual tracking.
MDNet \cite{Nam_2016_CVPR_MDNet} pretrains a shared CNN model and finetunes multiple domain-specific layers during online learning.
FCNT \cite{Wang_2015_ICCV_FCNT} employs a pretrained VGG-16 network \cite{simonyan2014very_VGG} and learns two additional head networks during visual tracking.
STCT \cite{Wang_2016_CVPR_STCT} exploits ensemble learning to utilize different CNN feature channels.
CREST \cite{Song_2017_ICCV_CREST} reformulates correlation filter as a network layer with residual learning.
DSLT \cite{Lu_2018_ECCV_DSLT} designs a shrinkage loss to improve deep regression tracking.

Recently, ATOM \cite{Danelljan_2019_CVPR_ATOM} designs the overlap maximization based architecture to predict the intersection over union (IoU) overlap between the target and the proposal boxes.
As a modification, DiMP \cite{Bhat_2019_ICCV_DiMP} replaces the modulation module with a parametric optimizer to further improve performance.
Note that, both ATOM and DiMP handle prediction modeling by leveraging the IoU-Net \cite{Jiang_2018_ECCV_IoU-Net}, which still belongs to the anchor-based detection paradigm.
In contrast, the prediction modeling in our SiamKPN is fully based on heatmap regressions for center point, target size and offsets estimation.

\begin{figure*}[t]
\begin{center}
\label{FigMainFramework}
  \includegraphics[width=1.0\textwidth]{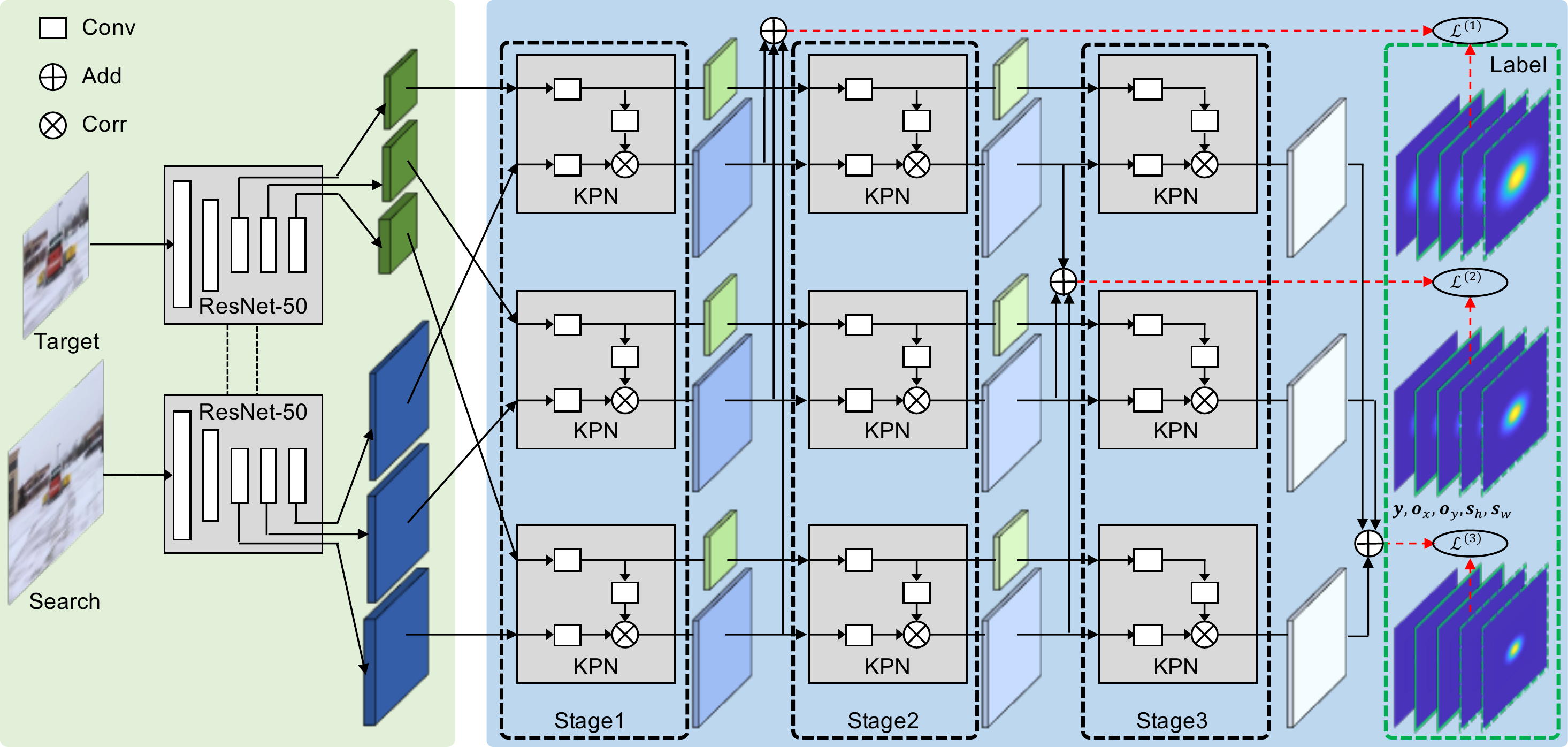}
\end{center}
\caption{Framework of SiamKPN. Given a pair of target and search image (left panel), we obtain convolutional feature maps from three different layers of a modified ResNet-50 Siamese backbone.
Each pair of target and search feature maps are further processed by several convolutions and a depth-wise cross-correlation in the cascade KPN head (right panel).
The label heatmaps are defined with different variances to enforce different strength of supervision over the output heatmap of each stage.}
\end{figure*}

\section{Model Representation}
In this section, we first present the basic building blocks of SiamKPN.
Then we detail the cascade heatmap scheme and illustrate its effect on coarse-to-fine prediction.
Figure \textcolor{red}{2} presents the whole framework of SiamKPN which consists of a Siamese backbone for feature learning, and a cascade of compact KPN heads for prediction modeling.

\subsection{Siamese Backbone}
Similar to \cite{Li_2019_CVPR_SiamRPNpp}, we employ a modified ResNet-50 to define our Siamese backbone network.
To make ResNet-50 suitable for our dense prediction task, we reduce the spatial stride to maintain more features and apply dilated convolutions to increase its receptive field.
In particular, the original spatial strides in conv$4\_1$ and conv$5\_1$ layers are converted to one thus yielding spatially larger feature maps.
Meanwhile, the original dilation rates in conv$4$ and conv$5$ blocks are changed to $2$ and $4$ to increase the receptive field.
Based on this modified ResNet-50 backbone subnetwork, we extract the outputs of conv$3\_4$, conv$4\_6$ and conv$5\_3$ layers as features, and apply $1\times 1$ convolutions to adjust channels before feeding them into the following head subnetwork. 

\subsection{Keypoint Prediction Head}
Figure \textcolor{red}{3} illustrates the architecture of the KPN head.
As show in the figure, it is constructed by three $3\times 3$ convolutions and one $5\times 5$ depth-wise cross-correlation.
More specifically, each KPN head involves the following flow of operations,
\begin{align}
\mathbf{\tilde y}^{(s)} &= \text{Corr}\left(\psi_{\mathbf{\hat y}}^{(s)},
    \text{Conv}\left(\psi_{\mathbf x}^{(s)}, \mathbf w_a^{(s)}\right)\right), \label{EqnPredHeat}\\
\psi_{\mathbf x}^{(s)} &= \text{Conv}\left(\mathbf{x}^{(s-1)}, \mathbf w_t^{(s)}\right), \\
\psi_{\mathbf{\tilde y}}^{(s)} &= \text{Conv}\left(\mathbf{\tilde y}^{(s-1)}, \mathbf w_s^{(s)}\right),
\end{align}
where Conv and Corr are abbreviations of convolution and cross-correlation.
Besides, $\mathbf w_t$ and $\mathbf w_s$ denote the parameters of the two convolutions for processing target and search feature maps respectively, $\mathbf w_a$ stands for that of an internal adjustment convolution.
We have made explicit the stage number $s$ to be consistent with the following notations
and $\{\mathbf x^{(0)}, \mathbf{\tilde y}^{(0)}\}$ are actually the output feature maps of the Siamese backbone for the target and search images.
Note that the search feature maps $\mathbf{\tilde y}^{(s)}$ are fed into the next stage while being further processed by another two convolutions to obtain the predicted heatmaps $\mathbf{\hat y}^{(s)}$.

Aside from the architecture, we would like to elaborate a little on the meanings of different channels in the predicted heatmaps.
In our implementation, three types of tasks are defined which include the center point, point offsets and target size estimation respectively.
In particular, we use one channel to handle the center point estimation task thus it represents the response map of the target location $\mathbf{\hat y}$.
Meanwhile, two offsets channels are utilized to address the discretization error due to stride resulting in $\{\mathbf{\hat o}_x, \mathbf{\hat o}_y\}$.
Besides, another two channels are employed to estimate the target size in terms of height and width $\{\mathbf{\hat s}_h, \mathbf{\hat s}_w\}$.

\begin{figure}[t]
\begin{center}
  \label{FigMainKPN}
  \includegraphics[width=.8\textwidth]{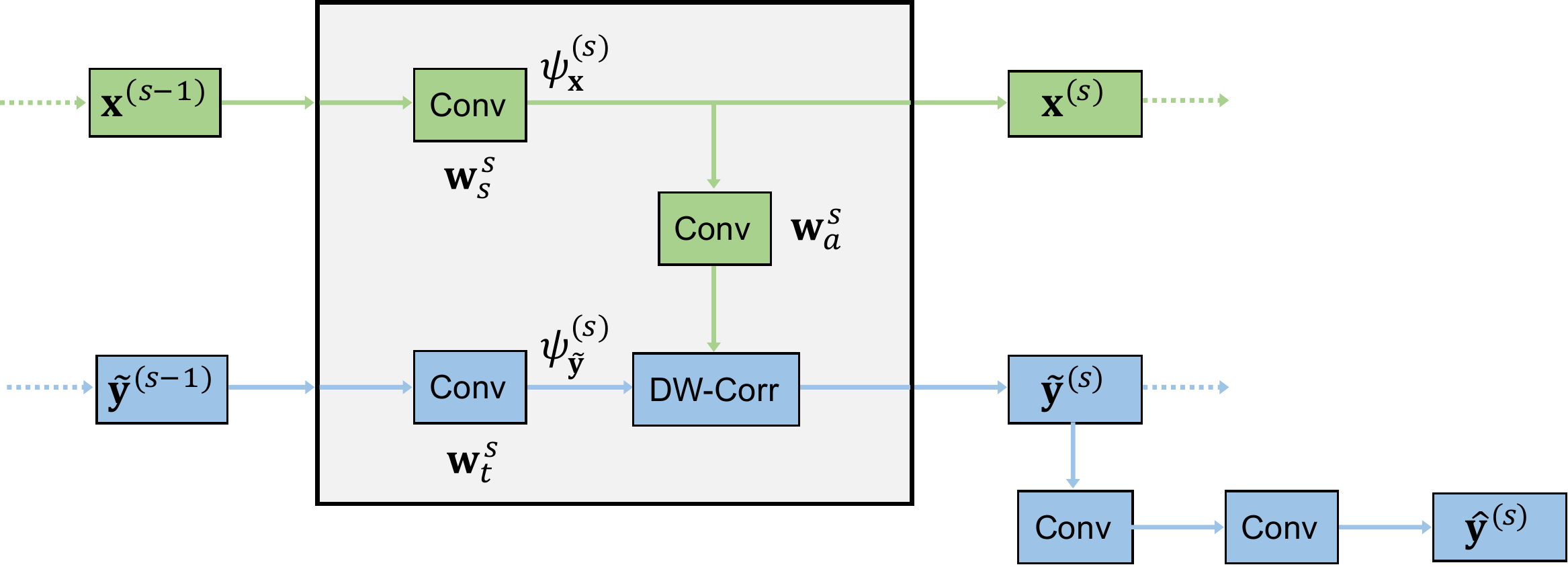}
\end{center}
\caption{Architecture of the keypoint prediction network. It consists of three convolutions and one depth-wise cross-correlation. Besides, the search feature maps are processed by another two convolutions to get the predicted heatmaps.}
\end{figure}

\subsection{Cascade Heatmap Supervision}
Unlike cascading the RPN head in \cite{Fan_2019_CVPR_C-RPN} which requires careful adjustments on the anchor scheme sequentially,
cascading the KPN head can be easily achieved by a direct repetitive stacking.
In addition, the whole architecture supports refinement along the cascade by successively shrinking the variance of the label heatmap.
To this end, we have the shrinking version of the Gaussian heatmap which is given as below,
\begin{align}
\mathbf y_{ij}^{(s)} &= \exp \left\{-\frac{(i-i_c)^2 + (j-j_c)^2}{2(\rho^{s-1}\sigma)^2} \right\}, \label{EqnLabHeat}
\end{align}
where $(i, j)$ denote the coordinates of an arbitrary point in the heatmap, $(i_c, j_c)$ represent the coordinates of a target center point, $s=1,2,3,\dots$ stands for the stage number, $\rho\in(0,1]$ controls the shrinkage strength of the variance $\sigma$ in the Gaussian function.
Hence, as the number of stage increases, the Gaussian heatmap becomes more peaked around the target center point.
In other words, the supervision signal is getting stricter along the cascade.

\begin{figure}[t]
\begin{center}
\subfigure[Fixed Variance]{
  \label{FigMainHeatmap1} 
  \includegraphics[width=.48\textwidth]{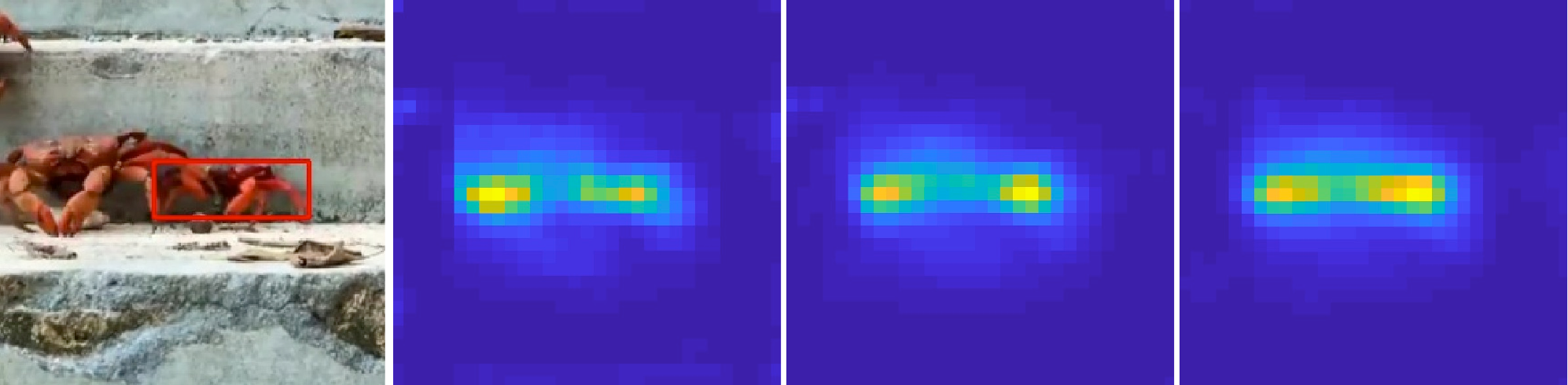}}
\subfigure[Shrinking Variance]{
  \label{FigMainHeatmap2}
  \includegraphics[width=.48\textwidth]{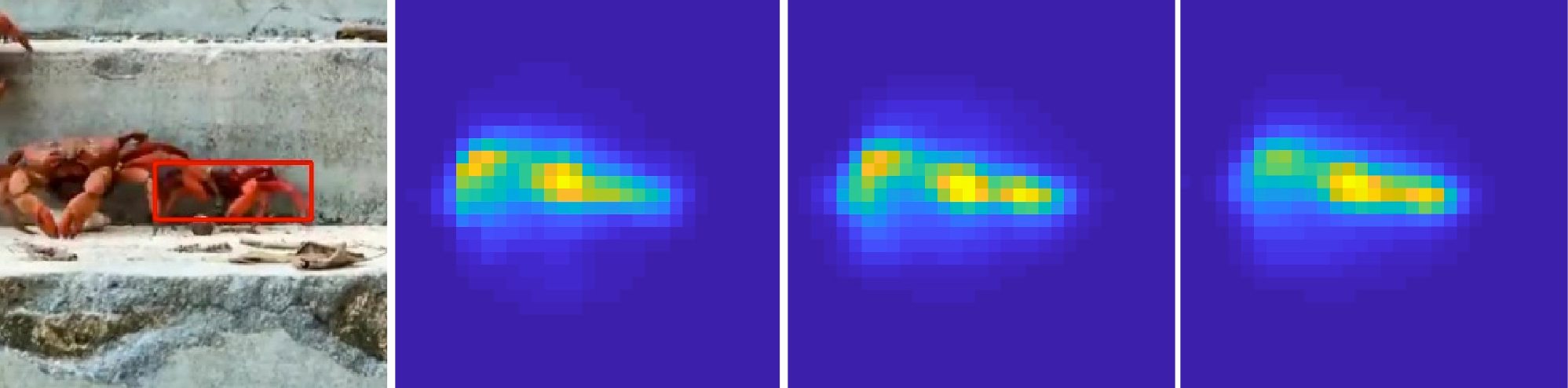}}
\end{center}
\caption{Visualization of the predicted heatmaps of the Crabs1 sequence from VOT2018 \cite{kristan2018vot} in successive stages of SiamKPN-3s with fixed or shrinkage variance. The right three columns correspond to the stage $1\sim 3$ respectively.}
\end{figure}

To illustrate the effect of our proposed scheme, we compare it with a naive stacking with fixed variance.
Figures \ref{FigMainHeatmap1} and \ref{FigMainHeatmap2} give an example of the sequentially evolving heatmaps of the two strategies.
As shown in the figures, the stacking with fixed variance can slightly help concentrate the predicted heatmap to the target, however it also strengthens the heatmap scores for the similar distractor.
In contrast, the shrinking variance scheme can well enhance the heatmap scores in the target and reduce that in the background, thus easier to distinguish the target from the similar distractor as the cascade gets deeper.

It is worth mentioning that, the above scheme can also be applied to point offsets and target size estimation.
In particular, the point offsets label $\{\mathbf o_x, \mathbf o_y\}$ are calculated as the discretization gap between the accurate position and a stride-clamped version, i.e., $\{\frac im - \lfloor\frac im\rfloor, \frac jm - \lfloor\frac jm\rfloor\}$ for position $(i,j)$ and stride $m$.
Besides, we define the target size label $\{\mathbf s_h, \mathbf s_w\}$ as the ground-truth height and width $\{h, w\}$ around the center point and zero otherwise.

\section{Algorithms}
In this section, we present the training and tracking algorithms for SiamKPN.
Generally speaking, the training of SiamKPN is done in an end-to-end manner with intermediate supervisions,
while the SiamKPN tracker is utilized without online adaptation to achieve high accuracy and robustness running in real time. 

\subsection{Offline Training}
During the offline training phase, we prepare the label heatmaps by setting $\rho=0.9$ and $\sigma=31/16$ considering that the search image feature map is in size $31\times 31$.
Similar to \cite{Law_2018_ECCV_CornerNet,Zhou_2019_arxiv_CenterNet}, we train each KPN head with a multi-task loss.
In particular, the keypoint estimation channel is optimized by a weight-balanced version of the focal loss \cite{lin2017focal},
while the offsets and target size estimation channels are supervised via a smoothed $\ell_1$ loss.

More specifically, by using Equations (\ref{EqnPredHeat}) and (\ref{EqnLabHeat}), the keypoint estimation loss between the predicted and label heatmap in our framework is defined as below
\begin{align}
\mathcal L_\text{kpt} =& -(1-\gamma)\sum_{i,j} \mathbb I[\mathbf y_{ij}=1](1-\mathbf{\hat y}_{ij})^\alpha\log\mathbf{\hat y}_{ij} \nonumber\\
&- \gamma\sum_{i,j} (1-\mathbf y_{ij})^\beta(\mathbf{\hat y}_{ij})^\alpha\log(1-\mathbf{\hat y}_{ij}),
\end{align}
where $\mathbb I[\cdot]$ denotes the indicator function.
Note that we have omitted the stage number $s$ for notational simplicity.
We set the focal loss hyper-parameters as $\alpha=2$, $\beta=4$ and $\gamma=0.05$ in all our experiments.

Besides, the offsets and target size estimation losses are defined as follows
\begin{align}
\mathcal L_\text{offs} &= \sum_{ij} \ell_1^\text{smooth}(\mathbf{\hat o}_{ij} - \mathbf o_{ij}), \\
\mathcal L_\text{size} &= \sum_{ij} \ell_1^\text{smooth}(\mathbf{\hat s}_{ij} - \mathbf s_{ij}), \\
\ell_1^\text{smooth}(x) &=
\left\{\begin{array}{ll}
0.5x^2 & \text{if } |x| < 1;\\
|x| - 0.5 & \text{otherwise}.
\end{array}\right.
\end{align}

By combining all the losses across different stages, we obtain the overall training objective as
\begin{align}
\mathcal L = \sum_s \mathcal L_\text{kpt}^{(s)} + \lambda_1\mathcal L_\text{offs}^{(s)} + \lambda_2\mathcal L_\text{size}^{(s)},
\end{align}
where $\lambda_1$ and $\lambda_2$ trade-off the balance among the keypoint, offsets and target size estimation.
We set $\lambda_1=1$ and $\lambda_2=0.05$ through all our experiments.

\subsection{Online Tracking}
For the online tracking phase, we first crop around the target and resize it to $127\times 127$ given its bounding box of the first frame in a video sequence.
After going through the backbone and the head subnetworks, the target feature maps are adjusted to obtain several matching templates in size $5\times 5$ along the cascade and these templates are then kept unchanged during the whole tracking process. 
Given the predicted target location in the previous frame, we crop an approximately two-times larger image region centered on this location and resize it to $255\times 255$.
After applying the $5\times 5$ templates along the cascade, we use the outmost response maps to predict the target location in the current frame.

To get the predicted bounding box, the center point response map is first transformed to the range from 0 to 1 by applying the sigmoid function.
By thresholding, the points with higher scores are involved in the subsequent process while the remaining points are ignored.
Based on the selected points, the candidate bounding boxes are obtained by considering the corresponding point offsets and target size.
To fight against the bounding box changes in adjacent frames, we add penalties for changes in target scale and aspect ratio given by the below function
\begin{align}
\tau_\text{penalty} &= \exp\left\{k*\max(\frac{s_1}{s_2}, \frac{s_2}{s_1})*\max(\frac{r_1}{r_2}, \frac{r_2}{s_1})\right\},
\end{align}
where $k$ is a hyper-parameter, $\{s_1,s_2,r_1,r_2\}$ represent the target scales and aspect ratios of adjacent frames respectively.
The score of each point is multiplied by its penalty factor $\tau_\text{penalty}$ to get the penalized scores.
Based on the penalized scores, we introduce a Gaussian smooth function to suppress large displacements from the target to yield the final score for each candidate point.
To this end, the location with the highest score corresponds to the predicted center of the target.
As a post-processing step, the target size of the center point is a weighted average using two adjacent frames.
Therefore, the whole process involves three hyper parameters, namely the penalty coefficient, the Gaussian smoothing window coefficient and the target size smoothing coefficient.

\section{Experiments}
In this section, we evaluate SiamKPN on several visual object tracking benchmarks including
OTB-100 \cite{wu2015otb}, VOT2018 \cite{kristan2018vot}, LaSOT \cite{fan2019lasot} and GOT-10k \cite{huang2018got}.
SiamKPN is implemented in Python using PyTorch on GTX 1080Ti GPUs.

\subsection{Implementation details}

\textbf{Network Architecture.}
The modified ResNet-50 backbone outputs the conv$3\_4$, conv$4\_6$ and conv$5\_3$ layer as features.
Their channels are adjusted from 512, 1024 and 2048 to 256, then fed into the KPN cascade.
When passing through the cascade, the channels of the target and the search feature maps are maintained at 256, while the spatial sizes are kept as $15\times 15$ and $31\times 31$ respectively.
Inside of each KPN, the adjustment convolution only operates on the $5\times 5$ center locations of the $15\times 15$ target feature map.
Hence, the obtained $5\times 5$ feature map plays as convolutional kernels over the search feature map during the depth-wise cross correlation.
The output of each KPN is further processed by two convolutions to give the predicted heatmaps in 5 channels.
A layer-wise aggregation is employed to merge the three branches to produce the final prediction of each stage.

\textbf{Training Datasets.}
The training datasets consist of the train splits from Youtube-BB \cite{real2017youtube}, LaSOT \cite{fan2019lasot}, GOT-10k \cite{huang2018got} and COCO \cite{lin2014microsoft} datasets.
In particular, Youtube-BB provides amounts of sparse-labeled videos, LaSOT and GOT-10k provides frame-by-frame labeled videos, and COCO images are used to increase class diversity.
With the ratio 4:2:2:1, we sample 450000 target-search pairs similar to \cite{Li_2018_CVPR_SiamRPN} per epoch with several data augmentation operations including random shift, random scale change, random blur, random color jitter and negative samples \cite{Zhu_2018_ECCV_DaSiamRPN}.
Each target-search pair of images are in size $127\times 127$ and $255\times 255$ which follows the same dataset curation procedure in \cite{bertinetto2016fully_SiamFC}.

\textbf{Learning.}
SiamKPN is trained by the adaptive moment estimation (Adam) optimizer \cite{Kingma_2015_ICLR_Adam} over 5 GPUs with a total batch size of 80. 
The ResNet-50 backbone is pre-trained on ImageNet \cite{russakovsky2015imagenet}.
During fine-tuning, only the last three convolutional blocks of the backbone and the whole head subnetwork are trained.
In particular, for the head subnetwork, the learning rate is step-wisely decayed from 0.005 to 0.002 for the first 5 epochs and exponentially decayed from 0.002 to 0.0005 for the last 15 epochs.
As for the backbone, we set the learning rate to zero for the first 10 epochs and one tenth of that in the head subnetwork for the last 10 epochs.
Finally, we set the shrinking factor $\rho$ to 0.9 for the supervision.

\textbf{Inference.}
During tracking, we normalize the predicted center point heatmap of the final stage by sigmoid function.
Based on this score map, we select top-scored points by thresholding it over 0.15 and clamping the number of points in the range from 8 to 32.
As for the three hyper-parameters, a two-level grid search is employed to find the best configuration.
The first-level searches each hyper-parameter from 0 to 0.9 with an interval of 0.1 for all the models of the last 10 epochs.
Then, we apply the second-level search with an interval of 0.01 around the so-far best configuration.

\subsection{Results on OTB-100 \cite{wu2015otb}}
\begin{figure}[t]
\begin{center}
  \subfigure{
    \label{FigExpOtb100_1}
    \includegraphics[width=.35\textwidth]{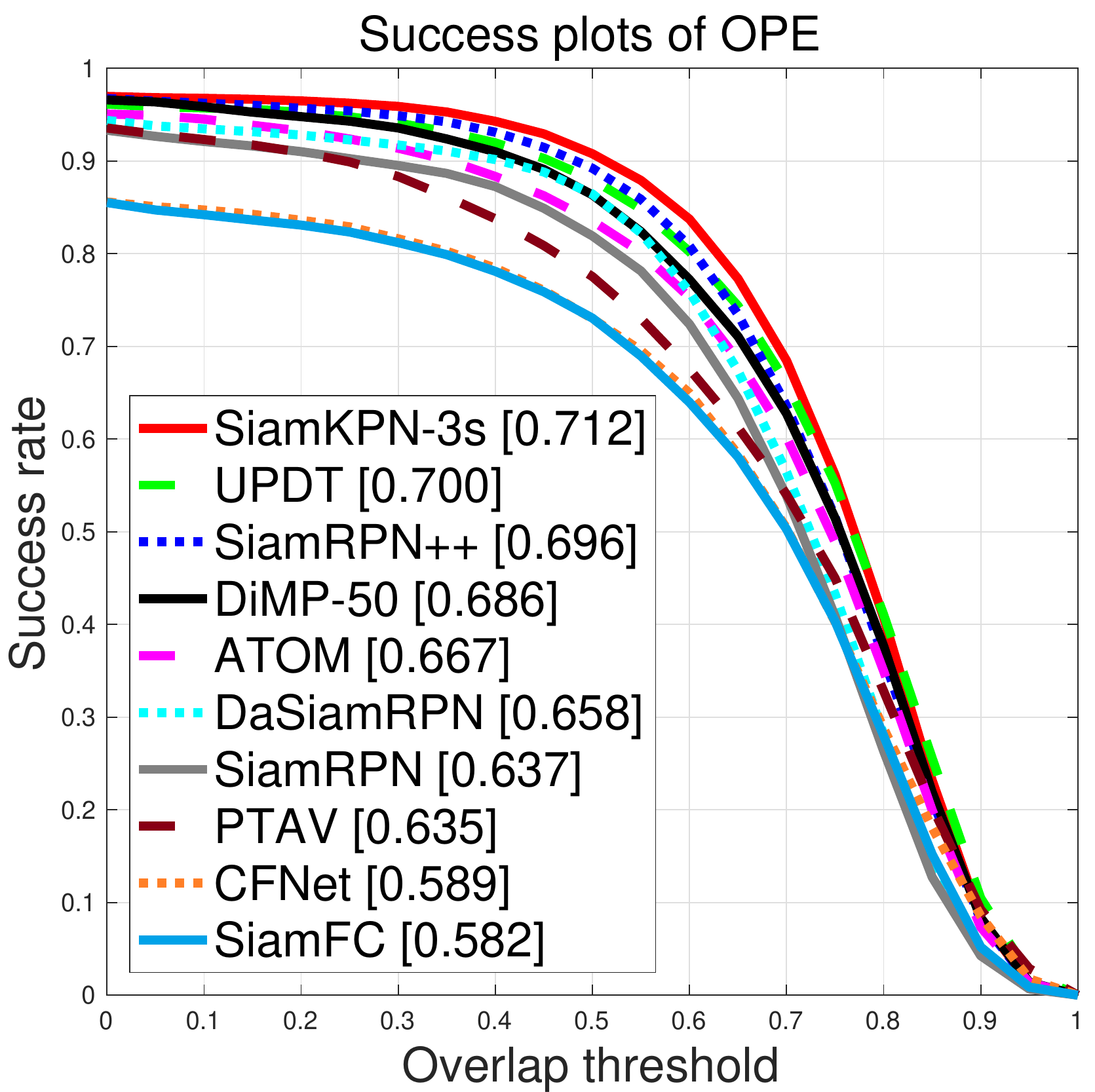}}
  \hspace{-1mm}
  \subfigure{
    \label{FigExpOtb100_2}
    \includegraphics[width=.35\textwidth]{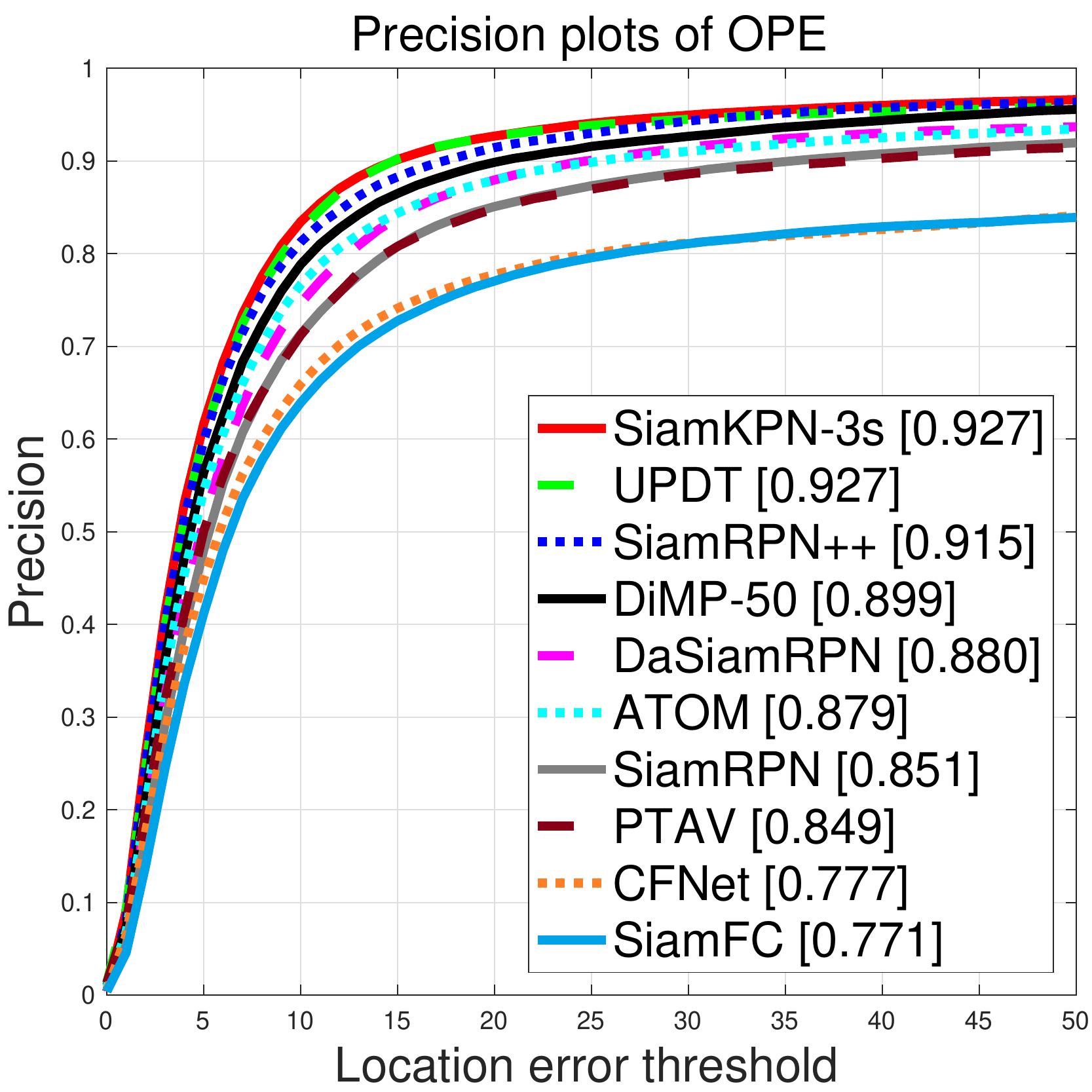}}
\end{center}
\caption{Comparisons with state-of-the-art trackers on the OTB-100 dataset. SiamKPN-3s achieves the best results on both success plot and precision plot. Best viewed in color.}
\end{figure}

OTB-100 contains 100 representative sequences and has eleven challenge attributes, including background clutters, scale variation, deformation and so on.
There are twenty grayscale sequences among them.
OTB uses one pass evaluation (OPE) to evaluate trackers with two metrics, precision and area under curve (AUC) of the success plot.
The precision plot illustrates the percentage of frames whose distance between the estimated location and ground truth is within the given threshold of 20 pixels.
The success plot is defined by the ratios of successful frames by varying the threshold from 0 to 1.

Figure \textcolor{red}{5} presents the comparison results between SiamKPN-3s and state-of-the-art trackers on OTB-100.
As shown in the figure, SiamKPN-3s ranks first both in success plot and precision plot and is ahead of the second-place tracker UPDT by 1.7\% in AUC score.
Besides, among all the Siamese trackers, SiamKPN-3s outperforms the previous best SiamRPN++ with relative gains of 2.3\% and 1.4\% in terms of success and precision respectively.
This validates that our cascade heatmap architecture can provide more accurate predictions.

\subsection{Results on VOT2018 \cite{kristan2018vot}}
\begin{table}[t]
\caption{State-of-the-art comparison on VOT2018 in terms of EAO, Accuracy and Robustness. The top-3 best results are highlighted by \textbf{bold}, \textit{italic} and \underline{underline}.}
\label{TabExpVot2018}
\centering
\resizebox{0.85\textwidth}{!}{
\begin{tabular}{|r|c|c|c||r|c|c|c|}\hline
~ & EAO $\uparrow$ & Accuracy $\uparrow$ & Robustness $\downarrow$ & ~ & EAO $\uparrow$ & Accuracy $\uparrow$ & Robustness $\downarrow$ \\\hline
SiamFC \cite{bertinetto2016fully_SiamFC} & 0.188 & 0.50 & 0.59 & UPDT \cite{Bhat_2018_ECCV_UPDT} & 0.378 & 0.536 & 0.184 \\
SiamRPN+ \cite{Zhang_2019_CVPR_CIR} & 0.30 & 0.52 & 0.41 & SiamRPN \cite{Li_2018_CVPR_SiamRPN} & 0.383 & 0.586 & 0.276 \\
DaSiamRPN \cite{Zhu_2018_ECCV_DaSiamRPN} & 0.326 & 0.569 & 0.337 & MFT \cite{kristan2018vot} & 0.385 & 0.505 & \textbf{0.140} \\
SPM \cite{Wang_2019_CVPR_SPM-Tracker} & 0.338 & 0.58 & 0.30 & LADCF \cite{xu2019learning_LADCF} & 0.389 & 0.503 & 0.159 \\
SiamMask \cite{Wang_2019_CVPR_SiamMask} & 0.347 & 0.602 & 0.288 & ATOM \cite{Danelljan_2019_CVPR_ATOM} & 0.401 & 0.590 & 0.204 \\
DRT \cite{Sun_2018_CVPR_DRT} & 0.356 & 0.519 & 0.201 & SiamRPN++ \cite{Li_2019_CVPR_SiamRPNpp} &  \underline{0.414} & \textit{0.600} & 0.234 \\
RCO \cite{kristan2018vot} & 0.376 & 0.507 & \underline{0.155} & DiMP-50 \cite{Bhat_2019_ICCV_DiMP} & \textbf{0.440} & \underline{0.597} & \textit{0.153} \\
~ & ~ & ~ & ~ & SiamKPN-3s & \textbf{0.440} & \textbf{0.606} & 0.192 \\\hline
\end{tabular}
}
\end{table}

\begin{figure}[t]
\begin{center}
  \label{FigExpVot2018}
  \includegraphics[width=.65\textwidth]{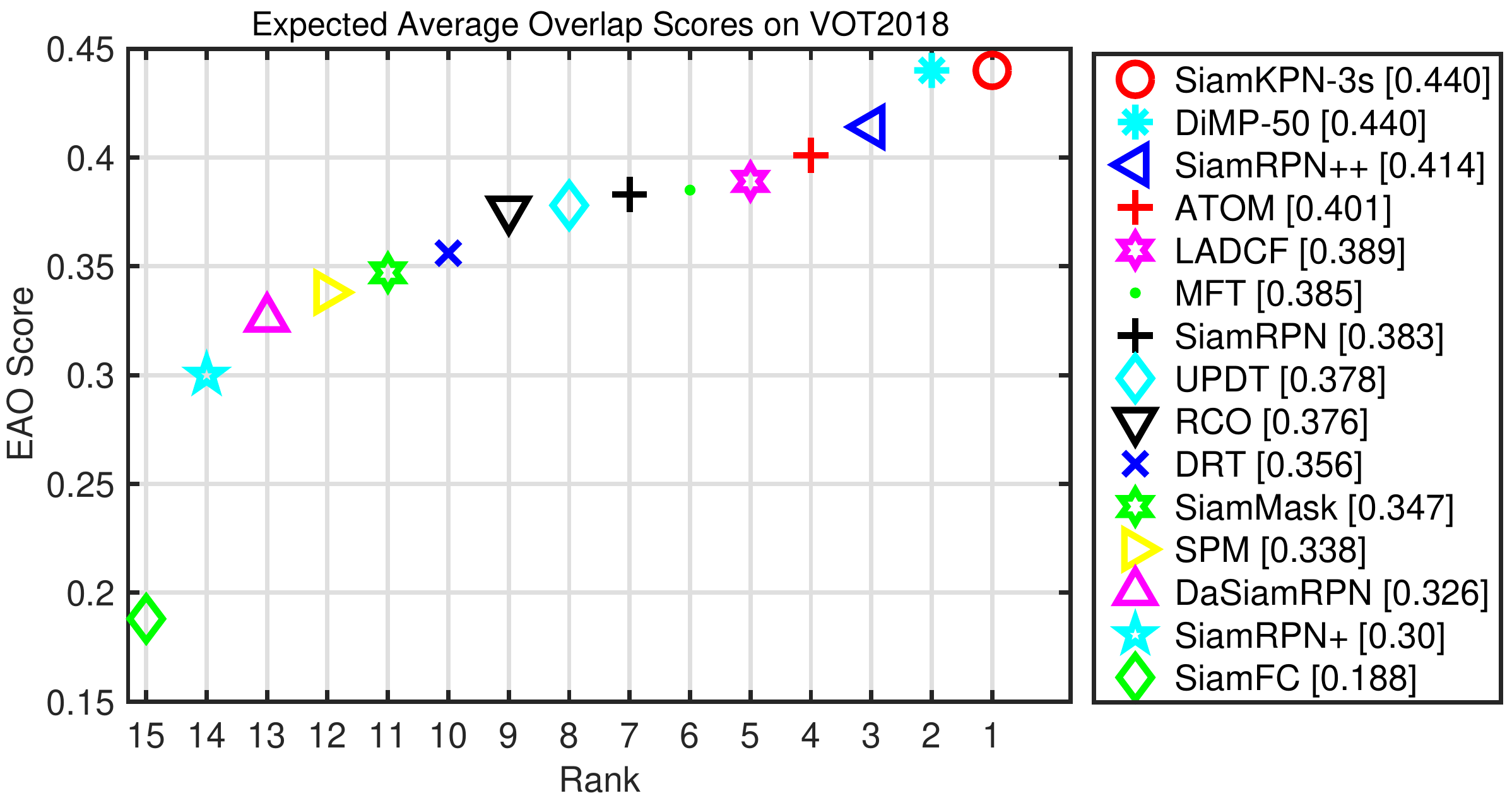}
\end{center}
\caption{Ranking of Expected Average Overlap results on the VOT2018 dataset. Best viewed in color.}
\end{figure}

\begin{figure}[t]
\begin{center}
  \label{FigExpVot2018speed}
  \includegraphics[width=.55\textwidth]{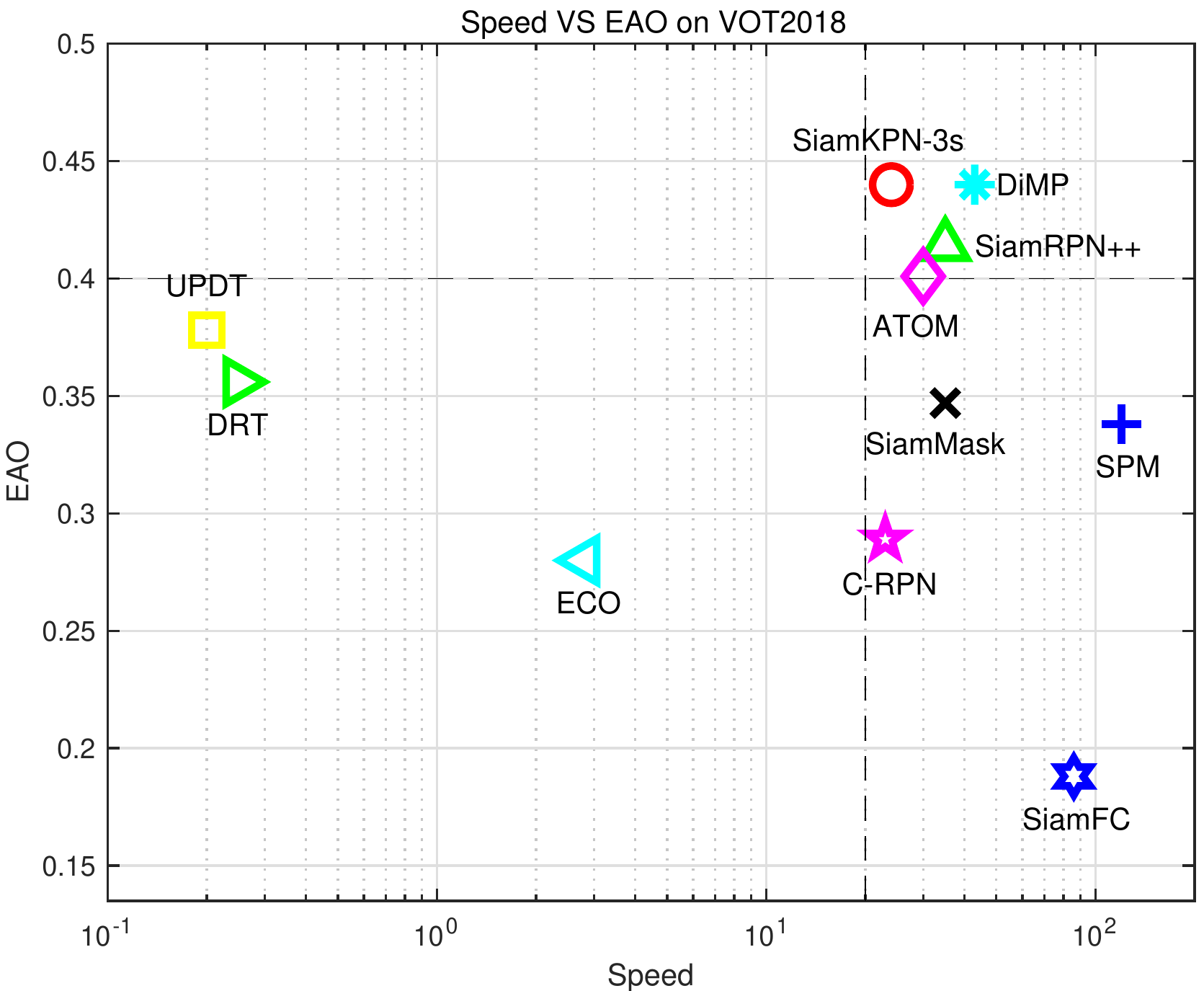}
\end{center}
\caption{EAO vs. Speed on the VOT2018 dataset. Best viewed in color.}
\end{figure}

VOT2018 contains 60 more challenging sequences.
Different from OPE test on OTB, VOT challenge reinitializes trackers at the failure frame.
It also has a burn-in period of ten frames, which means ten frames after initialization will be labeled as invalid for accuracy computation.
Besides, VOT uses accuracy (A), robustness (R), and expected average overlap (EAO) metrics to evaluate trackers.
Particularly, EAO score can comprehensively reflect accuracy and robustness.

We compare SiamKPN-3s with both Siamese trackers and DCF trackers on VOT2018.
As shown in Table \textcolor{red}{1} and Figure \textcolor{red}{6}, SiamKPN-3s ranks first according to EAO, surpassing SiamRPN++ and ATOM with a relative gain of 6.3\% and 9.7\% respectively. 
Without any online updating, our method achieves same EAO as DiMP and outperforms it on accuracy.
In addition, unlike other Siamese based trackers, SiamKPN-3s improves robustness in great margin.
The robustness score of SiamKPN-3s is better than all other Siamese trackers.
Even more, SiamKPN-3s has higher robustness score than ATOM, which is a discriminative tracker with online updating.
These results are reasonable since the variance-decay strategy can help SiamKPN-3s suppress distractors and improve robustness.

A comparison of quality and speed between SiamKPN-3s and state-of-the-art trackers on VOT2018 is shown in Figure \textcolor{red}{7}.
We visualize the Expected Average Overlap (EAO) with respect to the Frames-Per-Seconds (FPS). 
Note that the FPS axis is in the log scale and we set the real-time threshold at 20 FPS.
In the figure, the top-performed CF trackers run at slow speed while Siamese trackers run faster.
Our SiamKPN-3s gets best EAO score while runs at 24 FPS in real-time, achieving a better trade-off between performance and speed.


\subsection{Results on LaSOT \cite{fan2019lasot}}
\begin{figure*}[t]
\begin{center}
  \subfigure{
    \label{FigExpLasot_1}
    \includegraphics[width=.48\textwidth]{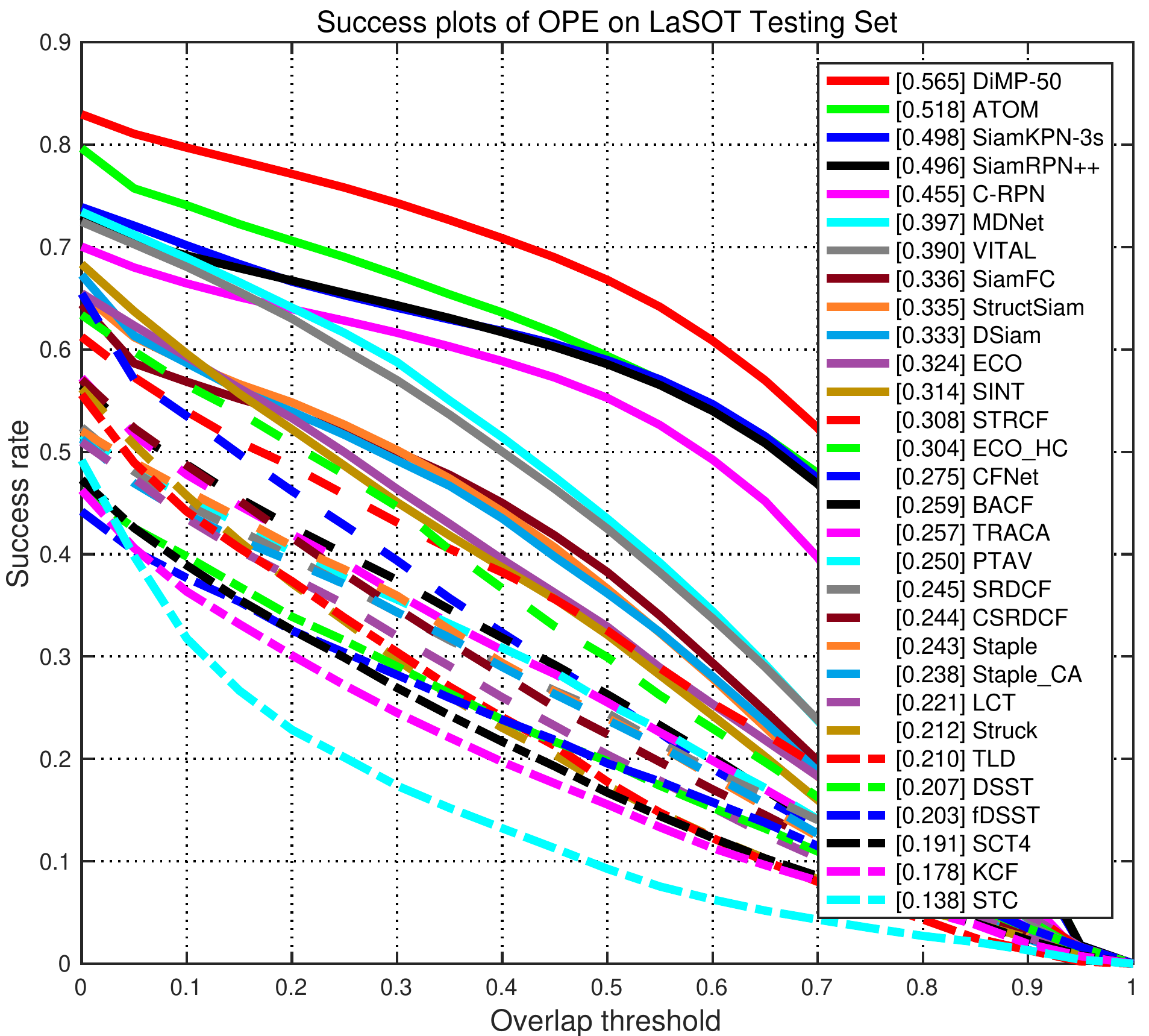}}
  \subfigure{
    \label{FigExpLasot_2}
    \includegraphics[width=.48\textwidth]{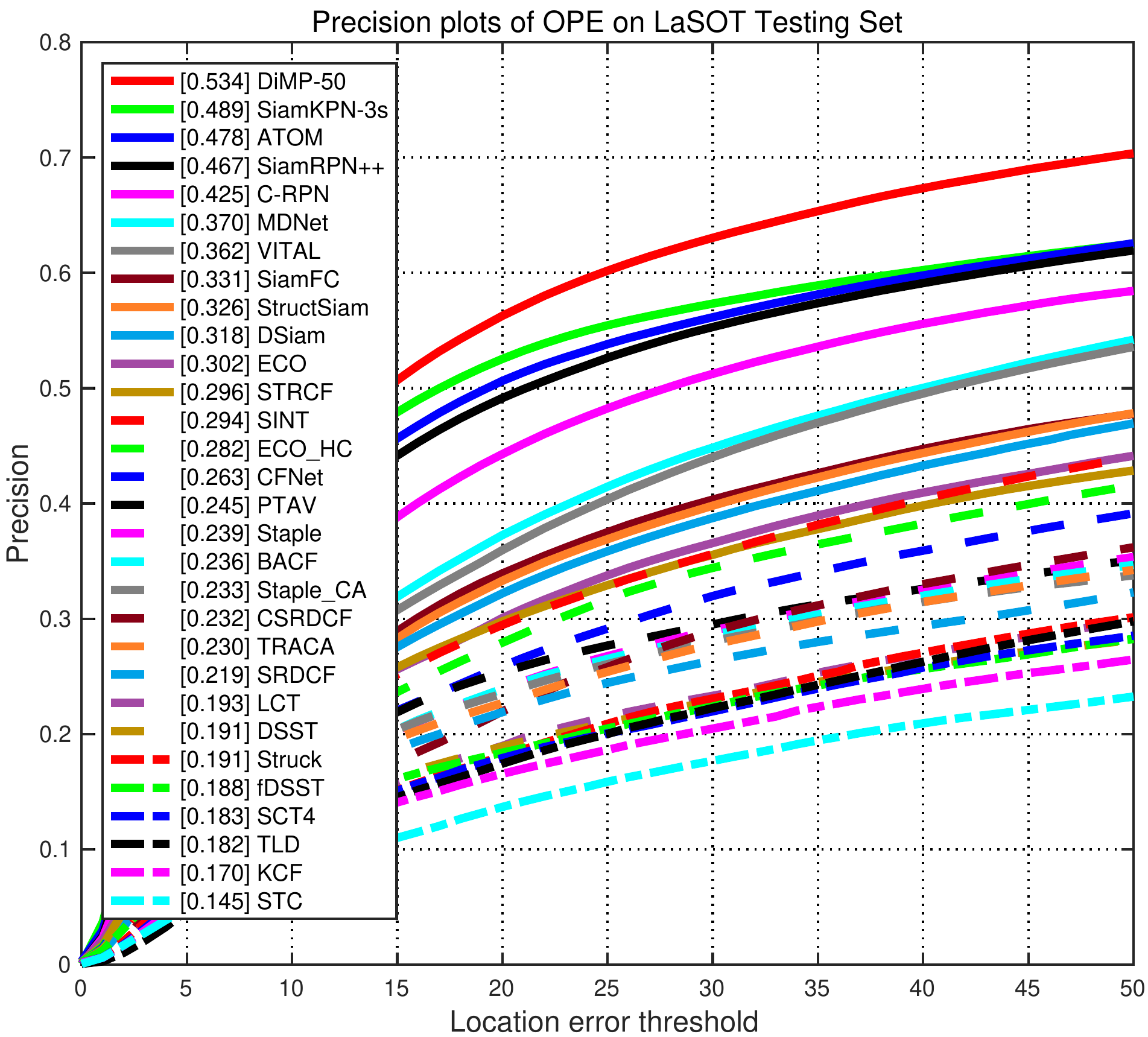}}
\end{center}
\caption{Experimental results on the LaSOT dataset. Best viewed in color.}
\end{figure*}

\begin{table}[t]
\caption{Experimental results on the LaSOT dataset. The top-3 best results are highlighted by \textbf{bold}, \textit{italic} and \underline{underline}.}
\label{TabExpLaSOT}
\centering
\resizebox{0.55\textwidth}{!}{
\begin{tabular}{|r|c|c||r|c|c|}\hline
~ & AUC $\uparrow$ & Pre $\uparrow$ & ~ & AUC $\uparrow$ & Pre $\uparrow$ \\\hline
KCF \cite{henriques2014high_KCF} & 0.178 & 0.170 & DSiam \cite{Guo_2017_ICCV_DSiam} & 0.333 & 0.318 \\
Staple \cite{Bertinetto_2016_CVPR_Staple} & 0.243 & 0.239 & StructSiam \cite{Zhang_2018_ECCV_StructSiam} & 0.335 & 0.326 \\
PTAV \cite{Fan_2017_ICCV_PTAV} & 0.250 & 0.245 & SiamFC \cite{bertinetto2016fully_SiamFC} & 0.336 & 0.331 \\
TRACA \cite{Choi_2018_CVPR_TRACA} & 0.257 & 0.230 & VITAL \cite{Song_2018_CVPR_VITAL} & 0.390 & 0.362 \\
BACF \cite{Galoogahi_2017_ICCV_BACF} & 0.259 & 0.236 & MDNet \cite{Nam_2016_CVPR_MDNet} & 0.397 & 0.370 \\
CFNet \cite{Valmadre_2017_CVPR_CFNet_SiamFCv2} & 0.275 & 0.263 & C-RPN \cite{Fan_2019_CVPR_C-RPN} & 0.455 & 0.425 \\
STRCF \cite{Li_2018_CVPR_STRCF} & 0.308 & 0.296 & SiamRPN++ \cite{Li_2019_CVPR_SiamRPNpp} & 0.496 & 0.467 \\
SINT \cite{Tao_2016_CVPR_SINT} & 0.314 & 0.294 & ATOM \cite{Danelljan_2019_CVPR_ATOM} & \textit{0.518} & \underline{0.478} \\
ECO \cite{Danelljan_2017_CVPR_ECO} & 0.324 & 0.302 & DiMP-50 \cite{Bhat_2019_ICCV_DiMP} & \textbf{0.564} &  \textbf{0.534}\\
~ & ~ & ~ & SiamKPN-3s & \underline{0.498} & \textit{0.489} \\\hline
\end{tabular}
}
\end{table}

To further validate the proposed framework on a larger and more challenging dataset, we conduct experiments on LaSOT.
The LaSOT dataset provides a large-scale, high-quality dense annotations with 1,400 videos in total and 280 videos in the testing set.
LaSOT has 70 categories objects with each containing twenty sequences, and the average video length is 2512 frames, which is useful to evaluate long-term trackers.
LaSOT adopts One-Pass Evaluation (OPE) test success and precision similar to OTB.
The precision is computed by comparing distance between predicted box and ground truth bounding box in pixels.
The success is computed as the Intersection Over Union (IoU) between predicted box and ground truth bounding box.

Table \textcolor{red}{2} and Figure \textcolor{red}{8} report the overall comparisons between our SiamKPN-3s tracker and other methods on LaSOT testing set.
Worth noting that the robustness measure is essential on LaSOT since zero-overlapping predictions will be included if a tracker loses the target in a long-term sequence.
From this point, trackers with online updating shall perform better on LaSOT.
Especially, DiMP ranks first in both success and precision.
However, among trackers without online updating, our SiamKPN-3s outperforms all other trackers, including SiamRPN++.
In particular, SiamKPN-3s achieves a larger precision score that is 4.7\% relatively higher than SiamRPN++.
Very unexpectedly, SiamKPN-3s even outperforms ATOM which is a recent tracker with online updating.
From the above comparison, SiamKPN-3s shows great potentiality for long-term tracking.

\subsection{Results on GOT-10k \cite{huang2018got}}
\begin{table*}[t]
\caption{Experimental results on the GOT-10k dataset. The top-3 best results are highlighted by \textbf{bold}, \textit{italic} and \underline{underline}.}
\label{TabExpGot10k}
\centering
\resizebox{\textwidth}{!}{
\begin{tabular}{|r|c|c|c|c|c |c|c|c|c|c |c|}\hline
~ & MDNet & CF2 & ECO & C-COT & GOTURN & SiamFC & SiamFCv2 & SiamRPN++ & ATOM & DiMP-50 & SiamKPN-3s \\
~ & \cite{Nam_2016_CVPR_MDNet}
& \cite{Ma_2015_ICCV_HCFT_CF2} & \cite{danelljan2016beyond_CCOT} & \cite{Danelljan_2017_CVPR_ECO}
& \cite{held2016learning_GOTURN} & \cite{bertinetto2016fully_SiamFC} & \cite{Valmadre_2017_CVPR_CFNet_SiamFCv2}
& \cite{Li_2019_CVPR_SiamRPNpp} & \cite{Danelljan_2019_CVPR_ATOM} & \cite{Bhat_2019_ICCV_DiMP} & ~ \\\hline
$\text{SR}_{0.50} \uparrow$ & 0.303 & 0.297 & 0.309 & 0.328 & 0.375 & 0.353 & 0.404 & \underline{0.615} & \textit{0.634} & \textbf{0.717} & 0.606 \\
$\text{SR}_{0.75} \uparrow$ & 0.099 & 0.088 & 0.111 & 0.107 & 0.124 & 0.098 & 0.144 & 0.329 & \textit{0.402} & \textbf{0.492} & \underline{0.362} \\
                    AO $\uparrow$ & 0.299 & 0.315 & 0.316 & 0.325 & 0.347 & 0.348 & 0.374 & 0.517 & \textit{0.556} & \textbf{0.611} & \underline{0.529} \\
\hline
\end{tabular}
}
\end{table*}

GOT-10k is also a large-scale dataset containing over 10000 video segments and has 180 test videos. 
The train and test splits have no overlap in object classes, thus overfitting on particular classes is avoided.
In addition, this benchmark requires all trackers to use the train split only for model training while external datasets are forbidden.
We strictly follow this protocol and retrain SiamKPN-3s solely on the train split of GOT-10k.
Table \textcolor{red}{3} shows the comparison results between SiamKPN-3s and other methods on the test split of GOT-10k.
Not astonishing, DiMP achieves the best performance since online updating is important for tracking unseen class object.
However, for trackers without online updating, our SiamKPN-3s achieves the best $\text{SR}_{0.75}$ and AO scores with relative gains of 10\% and 2.3\% over SiamRPN++.



\subsection{Ablation Study}
To investigate the impact of different components for our method, we conduct two ablation studies using OTB-100 and VOT2018.

\textbf{Numbers of Stages:} Table \textcolor{red}{4} presents the performance of SiamKPN by varying the number of stages.
With more stages, SiamKPN gathers more strength from the refinement process though in a diminishing manner of the performance gains.
It is also not astonishing that the tracking speed is decreased by increasing the number of stages.
However, the best performing SiamKPN-3s still runs at real-time speed with 24 FPS.
As a side observation, by recalling that in Figure \textcolor{red}{5} and Table \textcolor{red}{1}, one can find that the basic one-stage SiamKPN achieves comparable results with SiamRPN++ on the OTB-100 and VOT2018 benchmarks.

\begin{table}[t]
\caption{Effect on the number of stages in SiamKPN.}
\label{TabTestSetting}
\centering
\resizebox{0.47\textwidth}{!}{
\begin{tabular}{|r|c|c|c|c|}\hline
~ & One stage & Two stages & Three stages \\\hline
SUC on OTB-100 & 0.687 & 0.702 & 0.712 \\
PRE on OTB-100 & 0.906 & 0.916 & 0.927 \\\hline
EAO on VOT2018 & 0.413 & 0.428 & 0.440 \\
Speed on VOT2018 & 40 FPS & 32 FPS & 24 FPS \\
\hline
\end{tabular}
}
\end{table}

\textbf{Shrinking Variance:} During training, the shrinking variance in the heatmap supervision is important for guiding our framework to gradually refine the predictions.
To show its effectiveness, we compare the shrinking variance strategy using factor $\rho=0.9$ with a fixed variance strategy, i.e. $\rho=1$.
Table \textcolor{red}{5} shows the comparison results of SiamKPN-3s by using the two different strategies.
One can observe that, with shrinking variance, our tracker can predict more accurately on OTB-100.
This observation validates the importance of applying loose-to-strict supervision signals along the cascade.

\begin{table}[t]
\caption{Effect on using variance decay in SiamKPN-3s.}
\label{TabTestSetting}
\centering
\resizebox{0.42\textwidth}{!}{
\begin{tabular}{|r|c|c|c|c|}\hline
~ & w/o var decay & w/ var decay \\\hline
SUC on OTB-100 & 0.705 & 0.712 \\
PRE on OTB-100 & 0.916 & 0.927 \\
\hline
\end{tabular}
}
\end{table}


\section{Conclusion}
A Siamese keypoint prediction network has been developed for visual object tracking.
Although a lot of Siamese networks have been designed in the literature, there still lacks a high-performing scheme to address the challenges of the task.
The proposed SiamKPN model reduces this gap, by providing a cascade heatmap scheme to achieve both tracking accuracy and robustness.
Training SiamKPN is implemented with loose-to-strict supervisions and based on a multi-task loss in an end-to-end manner.
When applying SiamKPN for tracking, the predicted heatmap of successive stages has found to be gradually concentrated to the target while reduced to the distractors.
Based purely on offline training, SiamKPN performs favorably against state-of-the-art Siamese trackers and those methods with online learning modules, while running at real-time speed.

\clearpage

\bibliographystyle{splncs04}
\bibliography{SiamKPN}

\end{document}